\documentclass[letterpaper, 10 pt, conference]{ieeeconf}  

\IEEEoverridecommandlockouts                              

\usepackage{cite}
\usepackage{amsmath,amssymb,amsfonts}
\usepackage{algorithmic}
\usepackage{graphicx}
\usepackage{xcolor}

\usepackage{multicol}
\usepackage{float}
\usepackage{algorithm,algorithmic}
\usepackage{epstopdf}	
\usepackage{longtable}
\usepackage{caption}
\usepackage{subcaption}
\usepackage{comment}

\usepackage{tikz}
\usepackage{transparent}

\usepackage{mathtools}
\newtheorem{theorem}{Theorem}

\newtheorem{rem}{Remark}
\newtheorem{assumption}{Assumption}

\usepackage{multicol} 
 
\usepackage{nomencl} 
\makenomenclature
\renewcommand*\nompreamble{\begin{multicols}{2}}
\renewcommand*\nompostamble{\end{multicols}}

\newcommand{\figref}{Fig.~\ref}
\newcommand{\tabref}{Tab.~\ref}
\newcommand{\secref}{Sec.~\ref}

\begin{document}

\title{\LARGE \bf Towards Cooperative Flight Control Using Visual-Attention

\thanks{LY is supported by Knut and Alice Wallenberg Foundation. The research was partially sponsored by the United States Air Force Research Laboratory and the United States Air Force Artificial Intelligence Accelerator and was accomplished under Cooperative Agreement Number FA8750-19-2-1000. The views and conclusions contained in this document are those of the authors and should not be interpreted as representing the official policies, either expressed or implied, of the United States Air Force or the U.S. Government. The U.S. Government is authorized to reproduce and distribute reprints for Government purposes, notwithstanding any copyright notation herein. This work was further supported by The Boeing Company and the Office of Naval Research (ONR) Grant N00014-18-1-2830. Thank Paul Tylkin in the initial game setting and Hilal Hussain setting up the eye tracker.
}

}

\author{Lianhao Yin, Makram Chahine, Tsun-Hsuan Wang,  Tim Seyde, Chao Liu\\ Mathias Lechner, Ramin Hasani, Daniela Rus
\thanks{All the authors are with Computer Science and Artificial Intelligence Laboratory, Massachusetts Institute of Technology, Cambridge, MA, 02139 USA
    {\tt\small \{lianhao,tsunw, chahine, tseyde, cliu21, mlechner, rhasani, rus\} @mit.edu}}%
}

\maketitle

\begin{abstract}
The cooperation of a human pilot with an autonomous agent during flight control realizes parallel autonomy. We propose an air-guardian system that facilitates cooperation between a pilot with eye tracking and a parallel end-to-end neural control system. Our vision-based air-guardian system combines a causal continuous-depth neural network model with a cooperation layer to enable parallel autonomy between a pilot and a control system based on perceived differences in their attention profiles. The attention profiles for neural networks are obtained by computing the networks' saliency maps (feature importance) through the VisualBackProp algorithm, while the attention profiles for humans are either obtained by eye tracking of human pilots or saliency maps of networks trained to imitate human pilots. When the attention profile of the pilot and guardian agents align, the pilot makes control decisions. Otherwise, the air-guardian makes interventions and takes over the control of the aircraft. We show that our attention-based air-guardian system can balance the trade-off between its level of involvement in the flight and the pilot's expertise and attention. The guardian system is particularly effective in situations where the pilot was distracted due to information overload. We demonstrate the effectiveness of our method for navigating flight scenarios in simulation with a fixed-wing aircraft and on hardware with a quadrotor platform. 

\end{abstract}



\section{Introduction} \label{se:introduction}
\noindent It is challenging for a pilot to control an aircraft in a cluttered environment with unpredictable obstacles resulting from the high information density. Depending on the pilot's level of expertise, the pilot can quickly lead to information overload and potentially induce mission-critical failure modes \cite{Deveans2009}. In this scenario, an intelligent agent that is trained to pilot aircraft autonomously in challenging environments can help to operate in parallel to the pilot with minimal intervention to the human maneuver, enabling air-guardian autonomy.

Considering the pilot navigating to the goal positions, for example, during landing, the overloaded information on the numerous monitors may distract the attention of the pilots, which leads to instability of the aircraft control. 
This is when the guardian agent can be switched on to safely navigate the aircraft with greater attention to the stable flight when the pilot is not focused on the aircraft control. Achieving high levels of autonomy increases the safety of aerial robots in their cooperation with humans and other intelligent agents \cite{Lee1992,sofge2019ai}. 
An air-guardian system driven by advanced machine learning algorithms can make necessary interventions when a pilot performs poorly and is distracted by overloaded information. In this way, it safely co-pilots the aircraft to its destination, serving as an autopilot system \cite{Chao2007,8122757,Jackson2020}. 

\begin{figure*}[t] 
	\begin{center}
	\graphicspath{{}{pic/}}
     \includegraphics[width=0.9\textwidth]{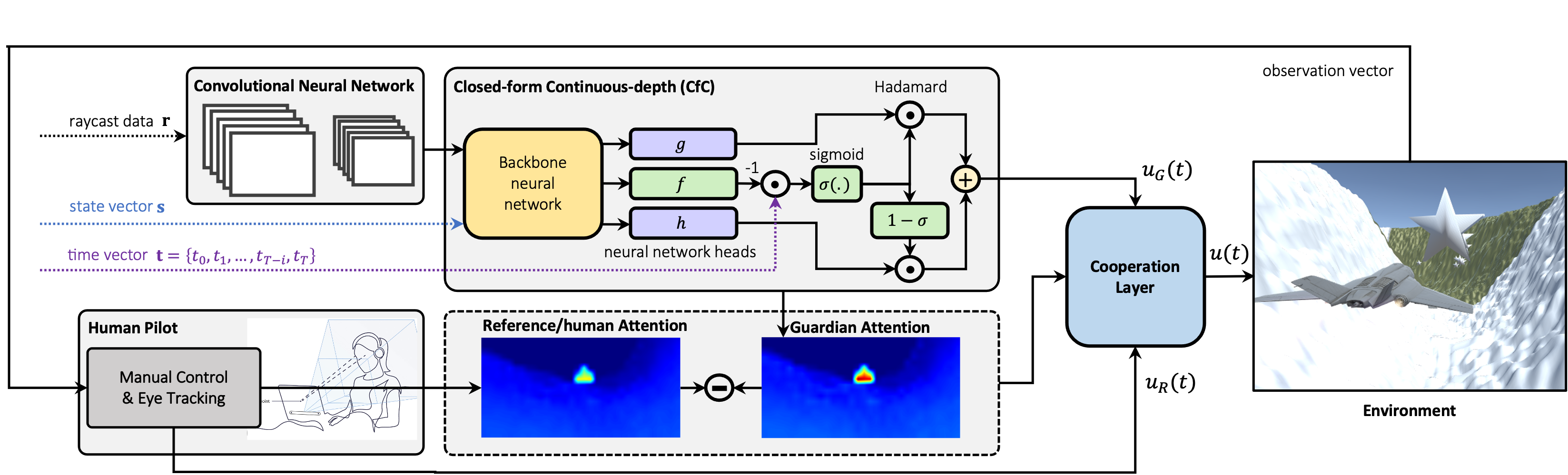}
	\caption{Cooperative control of a human pilot and our air-guardian. The air-guardian is an autonomous decision system that cooperates with the pilot to increase the safety of aircraft. A neural network takes the visual observation from the aircraft and predicts the control input $u_{G}$, while the pilot makes a decision and acts with $u_{R}$. A cooperative layer generates the applied control input $u$ using both $u_{G}$ and $u_{R}$ according to perceived attention mismatch.}
	\label{fig: front img}
	\vspace{-1mm}
	\end{center}
\end{figure*} 
In this paper, we design a novel air-guardian system based on visual attention, which can open new ways through how humans and autonomous agents cooperate. To this end, we first obtain a deep guardian agent via behavior cloning of expert data in a fixed-wing flight simulator. We compute the visual attention of the guardian agent by computing input importance metrics such as saliency maps and measure the attention of the human pilot using eye tracking. Our air-guardian agent then performs control through a cooperation layer which is a quadratic program that can regulate actions by comparing the attention maps of the two agents. By default, the pilot agent is controlling the aircraft; anytime the attention profile of the pilot and the guardian agent differ from each other, the guardian agent intervenes, yielding a safer maneuver of the aircraft with necessary intervention.

Compared to prior works and autopilot system \cite{Schwarting2018,Alshiekh2018}, which make interventions when the control of the pilot is outside of a safety region, our air-guardian system regulates interventions using a visual attention metric. As opposed to other baselines, this cooperation mechanism can potentially identify risky actions much earlier and develop more trust in the human pilot with its inherent interpretability. We are not aiming to replace all safety modules but to alert pilots and other safety modules and also cooperate with them in a more intelligent way.

\noindent \textbf{Contributions.} Our paper makes the following contributions:

\noindent 1) Introducing a novel cooperative flight control system using visual attention of network and eye tracking of human pilots. The cooperative network can assist a human pilot who is either less trained to achieve the target.
\noindent 2) Integrating a new class of continuous-time neural networks termed the Closed-form Continuous-depth (CfC) models \cite{vorbach2021causal} as the guardian to provide robust and causal attention maps. 
\noindent 3) Designing a new cooperative layer via a quadratic program that can switch the control between the guardian and the pilot agent. We deploy the cooperative layer at inference on the trained agent. Nevertheless, in principle, this layer is differentiable and could be trained as well.

\section{Related Works}
\noindent \textbf{Human-robot interaction.} Human-robotic cooperation has been widely used in aerospace \cite{Zhu2020}, vehicle \cite{Schwarting2017}, robotic assistant system \cite{Nikolaidis2017,Javdani2015,Gopinath2017} and social robots \cite{Carvalho2021}. 
A few challenges in human-robotic cooperation are explained here but are not limited to these. The goal of the human is unknown to robots \cite{Javdani2015}. One can either predict the goal of the user or formulate the problem as a partially observable Markov decision process. The human in the closed loop needs to retain trust in the robot \cite{Nikolaidis2017}, then the behavior of the robot and human can be mutually adapted to achieve a common goal. People also have different preferences over how much control robots should have over them. The end-user could also change the parameter of the closed-loop system and optimize how much control the human possesses \cite{Gopinath2017}. 

We can also understand human-robot cooperation as a safety system, meaning it will make interventions only when reaching critical safety limitations. An intuitive way for humans and robots to cooperate is via a safety system that intervenes in the humans' control commands or provides feedback when a potentially unsafe situation is detected \cite{Schwarting2018}. Another approach to combining the control commands of the human and the robot is to assign them a weighting in the form of a linear combination \cite{anderson2013intelligent,Schwarting2018}.

\noindent \textbf{Air-guardian.} An air guardian describes a system that monitors pilot control commands and switches to automatic control in certain situations \cite{Knapp2020}.
The switching of the control is triggered by measuring a risk \cite{Schwarting2018} or trust \cite{Lee1992} metric.
A model based on trust was suggested in \cite{Lee1992} to keep track of how frequently errors of human commands lead to unsafe scenarios. 
A risk metric of aircraft was proposed in \cite{Kim2017} that maps the state of an aircraft and the human's attention to a risk term. 
The state of the aircraft may comprise the spatial position, aircraft rotation angle, speed, accelerations, and distance to obstacles \cite{Wessendorp2021,Xu2015}. 
The main limitation of such an approach is that the state of an aircraft is not always accurate and available, e.g., distance to obstacles not accessible during visual navigation. 

\noindent \textbf{Visual Attention.} Saliency maps compute the visual attention of computer vision algorithms that process images, i.e., measuring the importance of the input pixels at inference.
Such \emph{attention maps} have been particularly important for analyzing which parts of an image black-box machine learning models focus their attention on \cite{Lecun2015,Xu2015_attention,Bojarski2016}.
For example, attention maps have been used to identify the input regions that have the highest influence on the network's output in end-to-end autonomous vehicle control~\cite{Yang2019, Kim2017}. The eye tracking system is a way to identify the attention of a human \cite{Aronson2018}. However, this paper uses the attention profile abstracted from a network that can imitate human behaviors instead of eye-gazing attention.

\noindent \textbf{Recurrent neural network.} Recurrent neural networks (RNNs) based on ordinary differential equations are known to learn causal relations well \cite{scholkopf2019causality}.
Liquid-time constant network \cite{Hasani2020}, a special type of such differential equations-based RNN, has been shown to be particularly successful at capturing causal dependencies in visual tasks.
For example, the attention map of this type of network aligns closer with that of human observers and is more robust to noise perturbations in vehicle and aircraft navigation problems~\cite{Lechner2020,vorbach2021causal,Chahine2023}. Moreover, RNNs in general and liquid-time constant networks specifically have been shown to be much more robust to noise than feedforward networks for the same task \cite{Lechner2020}. 
For our work, we adopt closed-form continuous-depth models (CfCs) \cite{hasani2021closed}, which is a computationally more efficient variant of the liquid-time constant networks~\cite{hasani2021liquid} that maintain some of its attractive salient learning capabilities, such as expressivity, causality, and robustness.

\section{Setup and Methodology} \label{section: Methodology}
In this section, we introduce our air-guardian concept. Figure \ref{fig: front img} shows an overview of our method. The air guardian agent, $f_{G}$, is composed of a stack of convolutional layers with closed-form continuous-depth (CfC) recurrent networks. The agent is trained via behavior cloning with expert data in a fixed-wing arcade environment \cite{PaulTylkinTsun-HsuanWangTimSeydeKylePalkoRossAllen2022}. The authors selected the best flying episode from a few hours' traces of experts flying in simulations and real drones. The reference mode $f_{R}$ is trained with the data from a specific pilot which can resemble pilots with different levels of expertise.

The goal is to safely navigate the aircraft in a low-altitude canyon environment in the presence of waypoints to be reached and obstacles to be avoided via the cooperation of the guardian and the pilot agent. The fixed-wing environment is to imitate the scenario where the pilot suffers from cognitive overload and fails to pay attention to critical environmental information. 

The cooperative layer computes the final output $u$ by combining the control input of the guardian network $u_{G}$ and the human pilot input $u_{R}$. This is regulated by the difference between the human attention map and the air-guardian attention map which is the weighted-center distance of the two attention maps. The weighted center is defined as $d_{c} = \sum{w_{i} x}/\sum{w_{i}}$, where the $w_{i}$ is the weight of each pixel and $x$ is the position of each pixel. 
The air-guardian attention map and the reference agent's attention map are derived by the VisualBackProp algorithm \cite{Bojarski2016} from the networks $f_{G}$ and $f_{R}$. Specifically, $f_{R}$ was either the saliency map of a network trained to imitate human pilots or the measured eye gazing of human pilots. In the following, we describe each piece in detail.

\subsection{The Guardian agent's neural model}
The air-guardian network takes the incoming images as the input of the convolution layers shown in Fig.~\ref{fig: front img}. The output of the convolution layers is flattened and fed to a dense layer, followed by a CfC network \cite{hasani2021closed}. The input $I$ of CfC networks is the latent variable from an upstream dense layer and the output $\mathbf{z}$ is the action for the aircraft. CfC networks are sequential decision-makers that, with the explicit model of timing behavior and varying time-constant, were shown to capture the underlying causal structure of a given task \cite{hasani2021closed}. This allows the extracted attention map to capture accurate information about the decision-making process. The state of the network is calculated by the liquid time-gating of $g$ and $h$ shown in Eq.~\ref{eq: CfC network}. 
Note that the gating is not only determined by the input $I$, state $x$, but also by the time $\mathbf{t}$, which makes this network a time-liquid recurrent network \cite{hasani2021liquid}. 
\begin{equation}\label{eq: CfC network}
\begin{aligned}
\mathbf{z}(t) = \sigma(-f (\mathbf{x}, \mathbf{I};\theta_{f})\mathbf{t}) \odot g(\mathbf{x},\mathbf{I};\theta_{g})
+ \\
(1-\sigma(-f (\mathbf{z}, \mathbf{I};\theta_{f})\mathbf{t}) ) \odot h(\mathbf{x}, \mathbf{I}; \theta_{h})
\end{aligned}
\end{equation}
The network $f$, $g$, and $h$ share the first few layers as a backbone and then branch out into three functions. The corresponding network parameters are $\theta_{f}$, $\theta_{g}$, and $\theta_{h}$. The time-decaying sigmoid function $\sigma$ stands for a gating mechanism that toggles between two continuous trajectories. $\odot$ is the Hadamard (element-wise) product.

\subsection{Visual attention map}
The attention maps of neural networks are calculated by the VisualBackProp algorithm~\cite{Bojarski2016} and that of humans are measured by eye tracking. For a convolutional neural network with weights and biases, we can think of a network flow from images to the output of the network \cite{Bojarski2016}. We denote the activation ${a_e}$ for one edge $e$ in the particular path $P$ in the general CNN. and $\mathcal{P}$ is a family of paths from image to output. From the value of pixel $X$, denote the backpropagation function \cite{Bojarski2016}: $\phi_{f}(X) = c \gamma(X)\sum_{P \in \mathcal{P} }  \prod_{e \in P}{a_e}$, where $c$ is the constant and $\gamma(X)$ is the value of pixel. Note that the subscript of the propagation function $\phi_{f}(X)$ is the network $f$, upon which the algorithm is used. The saliency maps corresponding to the reference and guardian networks are $\phi_{f_{R}}(X)$ and $\phi_{f_{G}}(X)$. 

\subsection{Cooperative network}
The cooperative network computes the final control $u$ by combining the pilot control $u_{R}$ and the air-guardian's $u_{G}$, taking into account the difference of the attention map of the pilot agent $\phi_{f_{R}}(X)$ and the guardian's $\phi_{f_{G}}(X)$. The weighted centers of the air-guardian saliency map and the pilot saliency map are denoted as $d_{\phi_{f_{R}}}$ and $d_{\phi_{f_{G}}}$, respectively. Denote the aircraft system as $\mathbf{x}_{k+1} = g(\mathbf{x}_k,\mathbf{u})$ with control input $\mathbf{u}$, state $\mathbf{x}$ and output $\mathbf{y}_k = q(\mathbf{x}_k)$. Moreover, let $I$ be the intervention of the guardian agent in the flight operation. The air-guardian will take control if $I = 1$ while the pilot agent will take control if $I = 0$. 
The switching of $I$ is regulated by the weighted-center distance of the attention maps (Eq.~\ref{eq: attention switch distance}).

\begin{equation} \label{eq: attention switch distance}
    I = 
\begin{cases}
      1 & \text{$ \|d_{\phi_{f_{R}}} - d_{\phi_{f_{G}}}  \| \leq d_f $}\\
      0 & \text{$ \|d_{\phi_{f_{R}}} - d_{\phi_{f_{G}}} \| > d_f $}
\end{cases}
\end{equation}

where, $d_f$ is the thresholds of the amplitude difference and weighted-center distance, respectively. 
The main reason for using weighted center distance as the main feature of the intervention is that the saliency map of the trained agent using CfC has a unique feature that highlights the goal positions, while another network such as LSTM has many highlights positions in the saliency map. If we use other networks as guardian agents or do not use saliency maps to abstract attention, one needs to use more features as criteria for intervention.

The optimization problem minimizes the difference between the control outputs from the reference and the air-guardian as follows: 
\begin{equation} 
\label{eq:MPC continous control}
\begin{aligned}
&\underset{\mathbf{u}_{k+1,k+2,...,k+H_p}}{\text{min}} && \sum_{i=k}^{i=k+Hp} \|\mathbf{u}_{i}-\mathbf{u}_{R_{i}}\|_{Q_{h}} + I \|\mathbf{u}_{i}-\mathbf{u}_{G_{i}}\|_{Q_{G}} \\
& \text{\boldmath{s}.\boldmath{t}.} && \mathbf{x}_{k+1} = g(\mathbf{x}_{k},\mathbf{u}_{k}) \\
& && \mathbf{y}_k = q(\mathbf{x}_k) \\
& && \mathbf{y}_k \in \mathbb{S}, \\
\end{aligned}
\end{equation}
where $H_{p}$ is the prediction horizon and $Q_{h}$, $Q_{G}$ are the weighting matrix. The solution of the optimization keeps the final control output $\mathbf{u}$ close to $\mathbf{u}_{R}$ when there is a small deviation in the attention profile of the reference and air-guardian agent. On the other hand, the solution keeps the final control $\mathbf{u}$ close to $\mathbf{u}_{G}$ when there is a large deviation. The weighting matrix also determines the portion of autonomy that the control system gives to the pilot. A higher weight on the $\mathbf{u}_{G}$ gives less controllability to the pilot. The robustness of the control system is discussed in Appendix. \ref{sec: appendix: robustness analysis}. 
In general, the intervention based on the attention profile influences more when the attention of the pilot is distracted, while the optimization in Eq.~\ref{eq:MPC continous control} keeps the system stable. This paper focuses more on discussing cooperation using attention profiles. 

The optimization problem Eq.~\ref{eq:MPC continous control} is differentiable and can be a layer of a neural network \cite{amos2017optnet}. This cooperation layer concludes our air-guardian concept.

\section{Human-in-the-loop Experiments with a simulated aircraft} 
\label{section: results and discussion}

\begin{figure}[ht] 
	\begin{center}
    \includegraphics[width=0.4\textwidth]{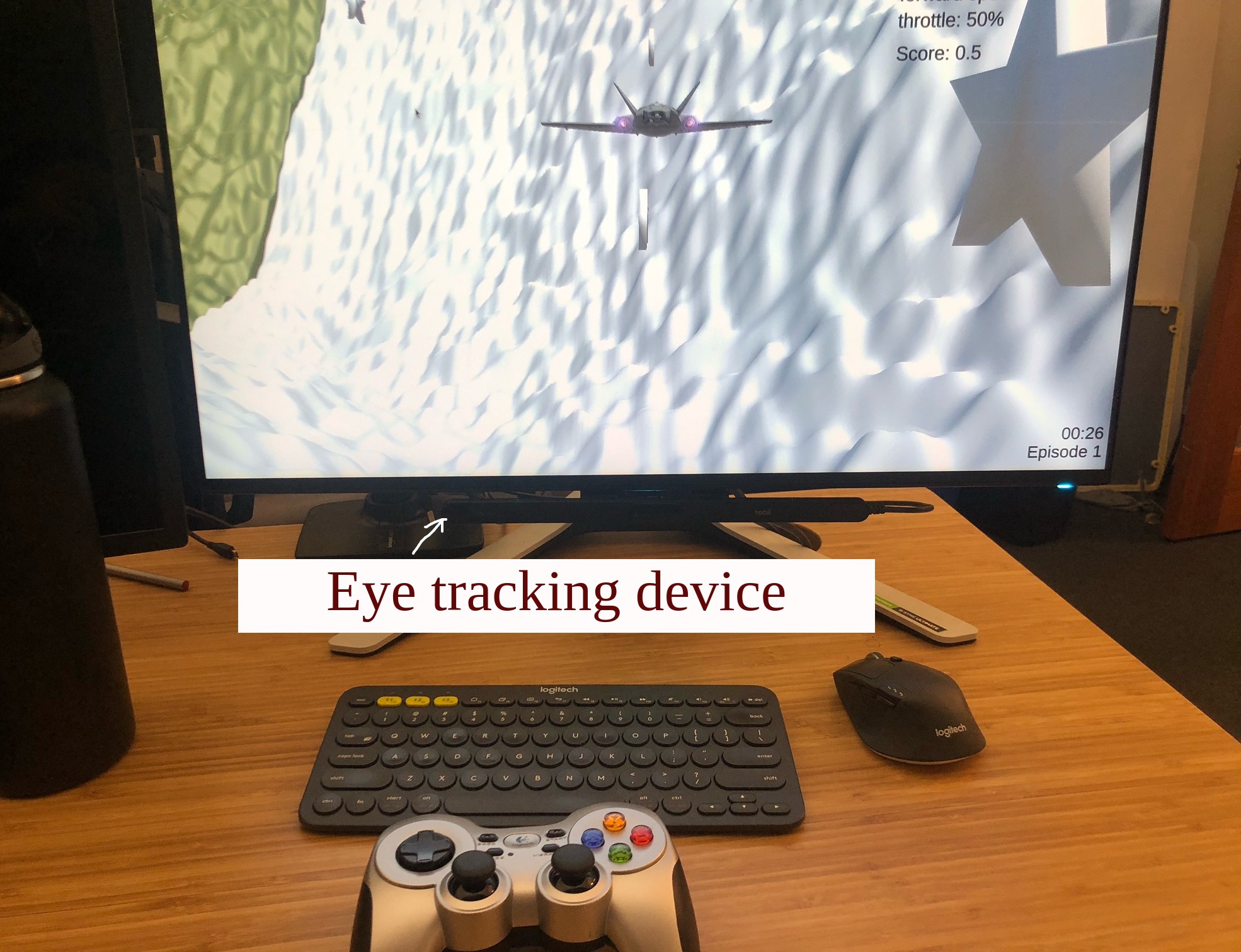}
     	\caption{The experimental setup with the eye tracking devices to get the attention profile of the pilot}
    	\label{fig:eye tracking device}
	\end{center}
\end{figure}

The experimental platform is an environment chosen from the autonomous flight arcade proposed in \cite{PaulTylkinTsun-HsuanWangTimSeydeKylePalkoRossAllen2022}. The attention profile of the guardian was derived from a trained agent and the attention profile of the human pilots was measured by the eye tracking device shown in Fig.~\ref{fig:eye tracking device}. 
The human testers were not trained in the designed flying task. The main reason is that the performance of the guardian network is not better than professional pilot due to limited training. The view of the human and guardian are exactly the same raw images, which makes sure the inputs to the human and guardian are the same. 
The aircraft is a fixed-wing. The basic aerodynamic effects are modeled. However, it does not consider the turbulence between the aircraft and its nearby objects. The observations are raycast which can be considered as Lidar signals. The reward comes from collecting stars at waypoints and the accumulated flying time. One gains a reward (0.5) at each catch of a waypoint and a reward (0.0001) for each additional flying step without crashing. The performance of the system is defined by the total reward during one episode of flying.

\subsection{Training setup}
The neural networks ($f_{G}$ defined in \secref{section: Methodology}) were trained using imitation learning. The training data contains 2 hours of expert trajectories which represents the optimal actions to gain reward by navigating along the canyon environment safely. 

\subsection{Results: guardian intervention}
In the experiment, a human pilot flew the aircraft using a joystick while eye-gazing positions were measured. In the experiment, the intervention profile for the guardian is the weighted center distance of guardian attention maps and the gaze of the pilot. The intervention process using control policy from Eq.~\ref{eq: attention switch distance} and Eq.~\ref{eq:MPC continous control} is illustrated in Fig.~\ref{fig: intervention process, exp2  }. The control was fully regulated by $u_{G}$ when the AI guardian was enabled and fully regulated by $u_{R}$ when the AI guardian was disabled. The attention difference threshold can be adjusted to have more strict or loose criteria for intervention. This criterion was tuned in the paper to balance the controllability of the human pilot and the total performance of the system. The control policy from Eq.~\ref{eq: attention switch distance} has a more cognitive motivation that when the difference of the attention focuses exceeds a threshold, the guardian takes intervention. Intuitively, when the attention of the pilot focuses on the wrong places, the guardian will take over control. This is especially important when pilots do not pay attention to the safety critical point in vision due to highly dense information from other aircraft operations. This resembles the cognitive overload situation, which is often one of the major factors leading to accidents in real-world aviation. 

\begin{figure}[th] 
	\begin{center}
    \includegraphics[width=0.45\textwidth]{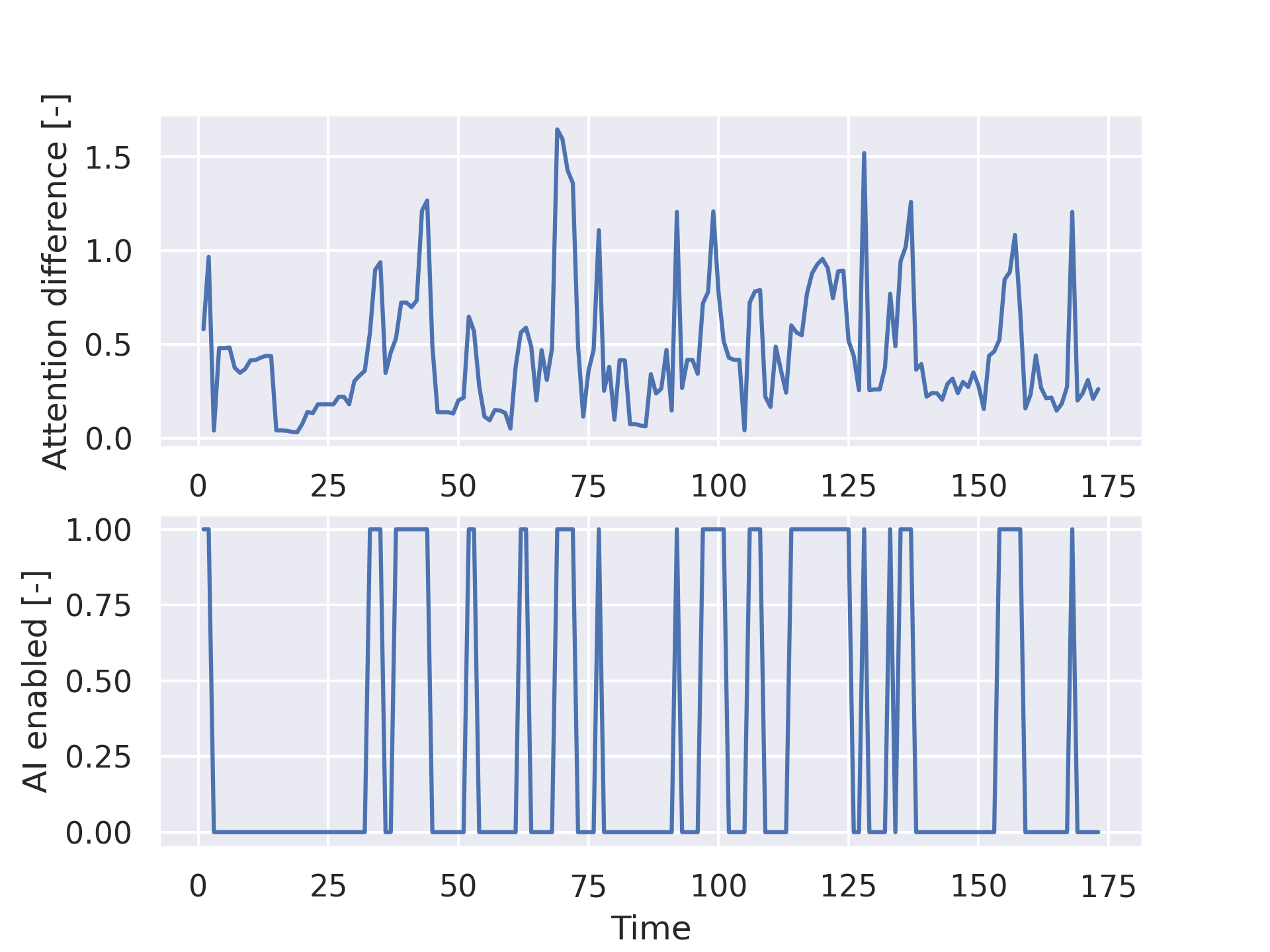}
     	\caption{The intervention process during a flight}
    	\label{fig: intervention process, exp2  }
	\end{center}
\end{figure}

\begin{figure}[ht] 
	\begin{center}
    \includegraphics[width=0.45\textwidth]{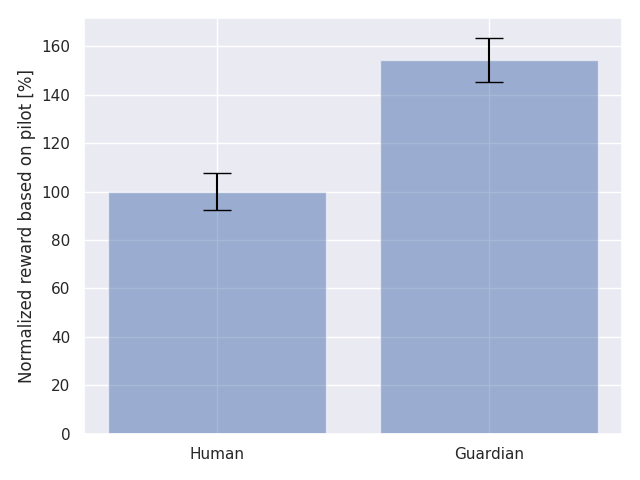}
     	\caption{Comparison of a human pilot with a guardian and human pilot}
    	\label{fig: comparison of guardian and human pilot}
	\end{center}
\end{figure}

In total 6 people participated in the experiments and each person played 10 rounds with a guardian agent and 10 rounds without a guardian agent. We observed that the collision rate of the reference pilot is 46\% and the collision rate with a guardian 23 \% over a constant length of flights with a p-value ($0.0004 < 0.05$). The difference is significant. The results show that the intervention policy defined in \eqref{eq: attention switch distance} can assist the human (Fig.~\ref{fig: comparison of guardian and human pilot}, Tab.~\ref{tb: experimental results in simulation}) when the pilots were distracted or not aligned with the expert attention. Each experiment conducted multiple runs, 20 each. The risk level is defined as the failure rate within the same flying distance in the simulation

\begin{table}[ht]
    \centering
    \caption{Experimental results in simulation}
    \begin{tabular}{c|c|c|c|c}
    \hline
            Name & Human & Guardian& Human& Guardian \\
          & Mean  & Mean & std & std  \\
    \hline
        reward & 0.31 & 0.49 & 0.38 & 0.47 \\
    \hline
         Risk level(\%) & 46     & 23 & 23 & 44 \\
    \hline
    \end{tabular}
    \label{tb: experimental results in simulation} 
\end{table}

\subsection{Analysis: robustness of the visual attention maps}
\begin{figure}[ht]
     \centering
     \begin{subfigure}[htbp]{0.20\textwidth}
         \centering
            \graphicspath{{}{pic/}}
		        \includegraphics[width=\textwidth]{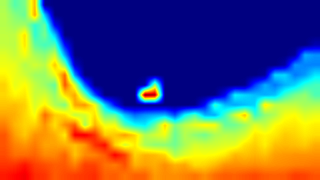}
         \caption{Attention map using CfC network.}
         \label{fig: VBP CfC }
     \end{subfigure}
     \hfill
     \begin{subfigure}[htbp]{0.20\textwidth}
         \centering
          \graphicspath{{}{pic/}}
		        \includegraphics[width=\textwidth]{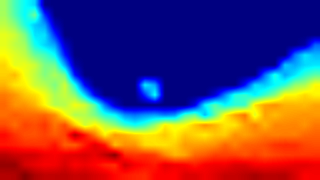}
         \caption{Attention map using LSTM.}
         \label{fig: VBP LSTM}
     \end{subfigure}
     \hfill
         \begin{subfigure}[htbp]{0.06\textwidth}
         \centering
          \graphicspath{{}{pic/}}
		    \includegraphics[width=\textwidth,angle=90,origin=c]{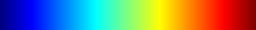}    %
     \end{subfigure}
    \hfill
     \begin{subfigure}[htbp]{0.20\textwidth}
         \centering
          \graphicspath{{}{pic/}}
		         \includegraphics[width=\textwidth]{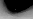}
         \caption{Ray cast images from the observation}
         \label{fig: ray cast}
     \end{subfigure}
     \hfill
        \begin{subfigure}[htbp]{0.20\textwidth}
         \centering
          \graphicspath{{}{pic/}}
		        \includegraphics[width=\textwidth]{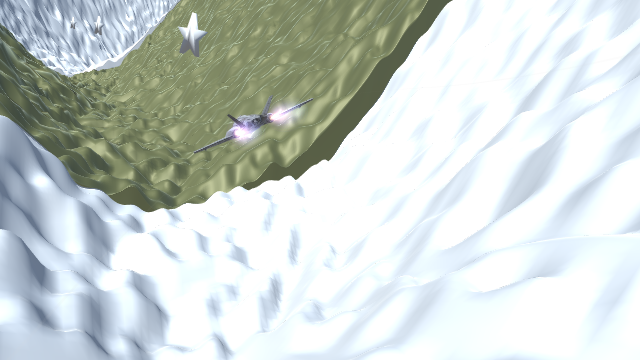}    
         \caption{Game image of current situation}
         \label{fig: game image}
     \end{subfigure}
    \hfill
         \begin{subfigure}[htbp]{0.06\textwidth}
         \centering
          \graphicspath{{}{pic/}}
		        \transparent{0.0}\includegraphics[width=\textwidth,angle=90,origin=c]{colorscale_jet.jpeg}    
     \end{subfigure}
     \hfill
     \caption{Comparison of attention maps (computed by VisualBackProp) between CfC and LSTM. The small dot is light blue and less visible and distinguishable from the rest of the figure (\figref{fig: VBP LSTM}). The small red dot is visible and distinguishable from the rest of the figure (\figref{fig: VBP LSTM}).}
    \label{fig:attention map comparison CfC and lstm}
\end{figure}

To justify the choice of the CfC model in our framework, we performed an additional experiment comparing the attention profile of a long short-term memory (LSTM) \cite{Hochreiter1997} with that of a CfC network. We train both modules with the same training data and process. One example comparison is shown in \figref{fig:attention map comparison CfC and lstm}. When the aircraft approaches the waypoint, the focus on the waypoint in the attention map generated from CfC (\figref{fig: VBP CfC }) is much more evident than that of LSTM (\figref{fig: VBP LSTM}). This enhances the robustness of the guardian agent as our concept is dependent on the attention profile of the pilot and guardian agent. 

\subsection{Analysis: interpretability}
\begin{figure}[h]
     \centering
     \begin{subfigure}[htbp]{0.20\textwidth}
         \centering
            \graphicspath{{}{pic/}}
		        \includegraphics[width=\textwidth]{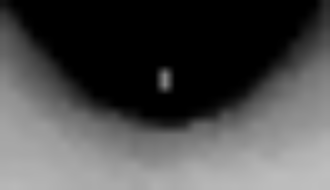}
         \caption{Raycast observation}
         \label{fig:interv obs}
     \end{subfigure}
     \hfill
     \begin{subfigure}[htbp]{0.20\textwidth}
         \centering
          \graphicspath{{}{pic/}}
		        \includegraphics[width=\textwidth]{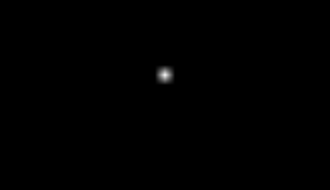}
         \caption{Eye attention}
         \label{fig:interv eye}
     \end{subfigure}
     \hfill
    \begin{subfigure}[htbp]{0.06\textwidth}
         \centering
          \graphicspath{{}{pic/}}
		        \transparent{0.0}\includegraphics[width=\textwidth,angle=90,origin=c]{colorscale_jet.jpeg}   
     \end{subfigure}
     \hfill
     \begin{subfigure}[htbp]{0.20\textwidth}
         \centering
          \graphicspath{{}{pic/}}
		         \includegraphics[width=\textwidth]{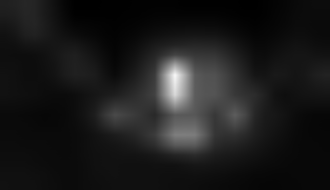}
         \caption{Attention profile and saliency map}
         \label{fig:interv saliency}
     \end{subfigure}
     \hfill
     \begin{subfigure}[htbp]{0.20\textwidth}
         \centering
          \graphicspath{{}{pic/}}
		         \includegraphics[width=\textwidth]{canyonrunpic/0680.png}
         \caption{The game view of such situation}
     \end{subfigure}
     \hfill
    \begin{subfigure}[htbp]{0.06\textwidth}
        \centering
          \graphicspath{{}{pic/}}
		        \transparent{0.0}\includegraphics[width=\textwidth,angle=90,origin=c]{colorscale_jet.jpeg}   
     \end{subfigure}
     \hfill
     
     \caption{The observation, saliency map of trained agent and human eye gazing point}
    \label{fig: intervention illustration }
\end{figure}

The attention map provides an intuitive representation of the key features the agent focuses on during its decision-making process since it highlights the saliency of key features in the environment. This makes potential mismatches in attention profiles and resulting interventions readily interpretable.
Specifically, the part of the attention map with a higher value has a stronger influence on the output action. When the waypoint showed up as the white dot in Fig.~\ref{fig:interv obs}, the attention intensity of the waypoint in the air-guardian's attention map was very strong shown in Fig.~\ref{fig:interv saliency} and the pilot was paying attention to that as well shown in the eye tracking (Fig.~\ref{fig:interv eye}). 

It is straightforward to interpret the difference in the attention maps of the guardian and the pilot based on the mismatch of the attention profiles.
When the difference in attention profile exceeds the threshold given in \eqref{eq: attention switch distance}, the air-guardian intervenes and takes dominant control of the aircraft. 
The human pilot has a lower performance to achieve a high reward in the designed mission and makes decision primarily based on environmental constraints. The guardian agent is able to negotiate a trade-off between safety and mission completion by distributing attention among the environmental constraints and waypoint objectives.

\section{Human-in-the-loop results on a real drone}
\begin{figure}[ht] 
	\begin{center}
	\graphicspath{{}{pic/}}
	\includegraphics[width=0.45\textwidth]{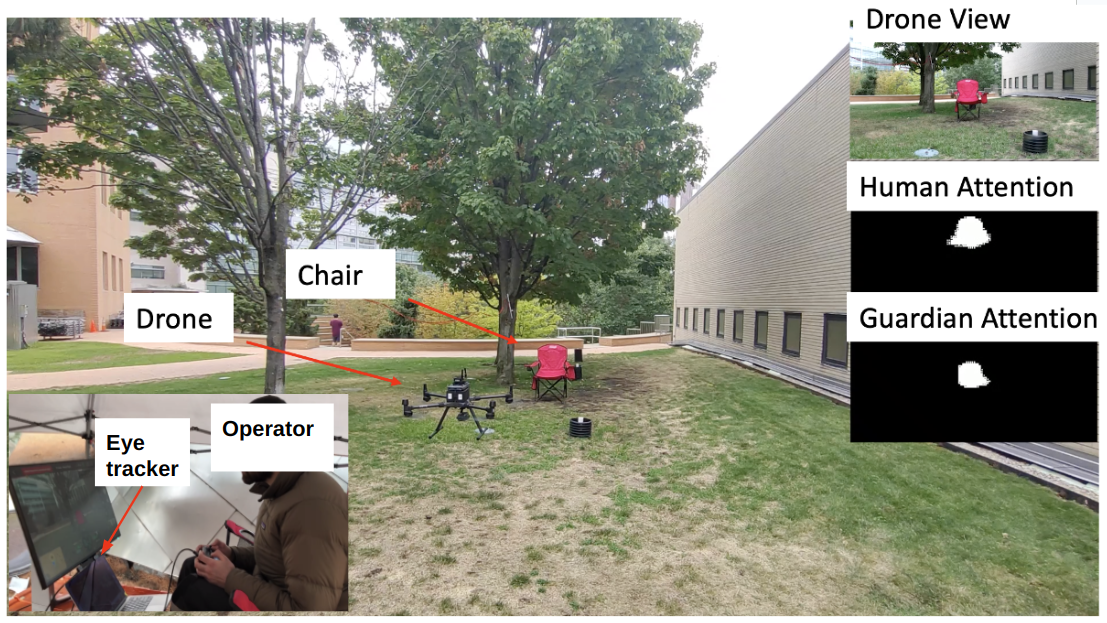}    
	\caption{The experiment consists of having a human pilot a quadrotor towards a target. Three subviews are drone's view, human attention and guardian attention}
	\label{fig: Field test setup}
	\end{center}
\end{figure}

The Human-in-the-loop experiment was defined as a task to pilot a quadrotor to its destination which is a red chair. The testing environment can be seen in Fig.~\ref{fig: Field test setup}

\subsection{Experimental settings}
The platform used is a DJI M300 RTK along with its onboard computer companion (DJI Manifold 2) that enables programmatic control of the drone with input images gathered from the gimbal stabilized Zenmuse Z30 camera.

The real drone experiments use RGB images instead of raycast used in the simulation as input to the neural network. The neural network for the real drone was trained using data collected in real experiments, and is able to accomplish the proposed task. The goal is to demonstrate the cooperative mechanism using attention in the real world. 
\subsection{Neural network structure and training}
 We pre-trained networks \cite{Chahine2023} for the task of flying towards the target from sequences of expert demonstrations using a Mean Squared Error (MSE) loss. The training data was collected in the woods. It contains roughly 100 expert demonstrations in multiple seasons. The year-round collection adds variance in the training data, with runs containing backgrounds rich in green leaves in the spring, others containing complex structures and a palette of dead leaves in the fall, and finally, data containing the striking contrast of dark, leafless trees against snow in the winter. In addition, closed-loop augmentation is performed by generating synthetic labeled data sequences from the original training set to enable the drone to recover from unseen situations.
All networks share a 5-layer CNN backbone taking a 144x256 RGB image as input, followed by an RNN head outputting the 4 dimensional control vector consisting of the translational velocities and yaw rate commands, fed into the quadrotor's lower level flight controller. Comparison of the input and output of the neural-network used in the simulation and the real drone are listed in Tab.~\ref{tab:comparison of network}. The RNN used in the experiment is a CfC architecture, which we take to be the guardian $f_{G}$ as defined in \secref{section: Methodology}.

\begin{table}[ht]
    \centering
    \caption{Neural network structure comparison}
    \begin{tabular}{c|c|c}
    \hline
         Drone experiments & Simulation & Real drone   \\
    \hline
         Observation & Raycast    & RGB  \\
         Guardian training & BC     & BC  \\
         CNN layer  & 3 & 5 \\
         Downstream network & CfC & CfC \\
         Number of control & 2  & 4 \\ 
    \hline
    \end{tabular}
    \label{tab:comparison of network}
\end{table}

The training data for the guardian are the whole expert data. We used the saliency map of a network that imitated a human pilot's behavior. 

\subsection{Experimental results}
In the test, we first verify the robustness of the system using the saliency map from a neural network imitating a human pilot for $f_{R}$ defined in \secref{section: Methodology} as the human pilot’s attention map. This network was trained using one pilot's trajectory. Then, we used gaze information as humans' attention measured by an eye-tracking system to verify the effectiveness of the guardian system. 

In the real world experiment, the intervention criterion is the same as that used in the simulation (the distance between the weighted center of human and guardian attention defined in Eq.~\ref{eq: attention switch distance}). 
The experiment consists of having a human pilot the quadrotor towards a target (red camping chair) with/without the help of the attention map-based guardian (\figref{fig: Field test setup}). In total, six people participated in the experiment, each repeating the task 16 times from four different starting points. Among the 16 flights, 8 are performed solely by the human, and the other 8 are performed with the guardian system on. 
\begin{figure}[h] 
	\begin{center}
	\graphicspath{{}{pic/}}
	\includegraphics[width=0.45\textwidth]{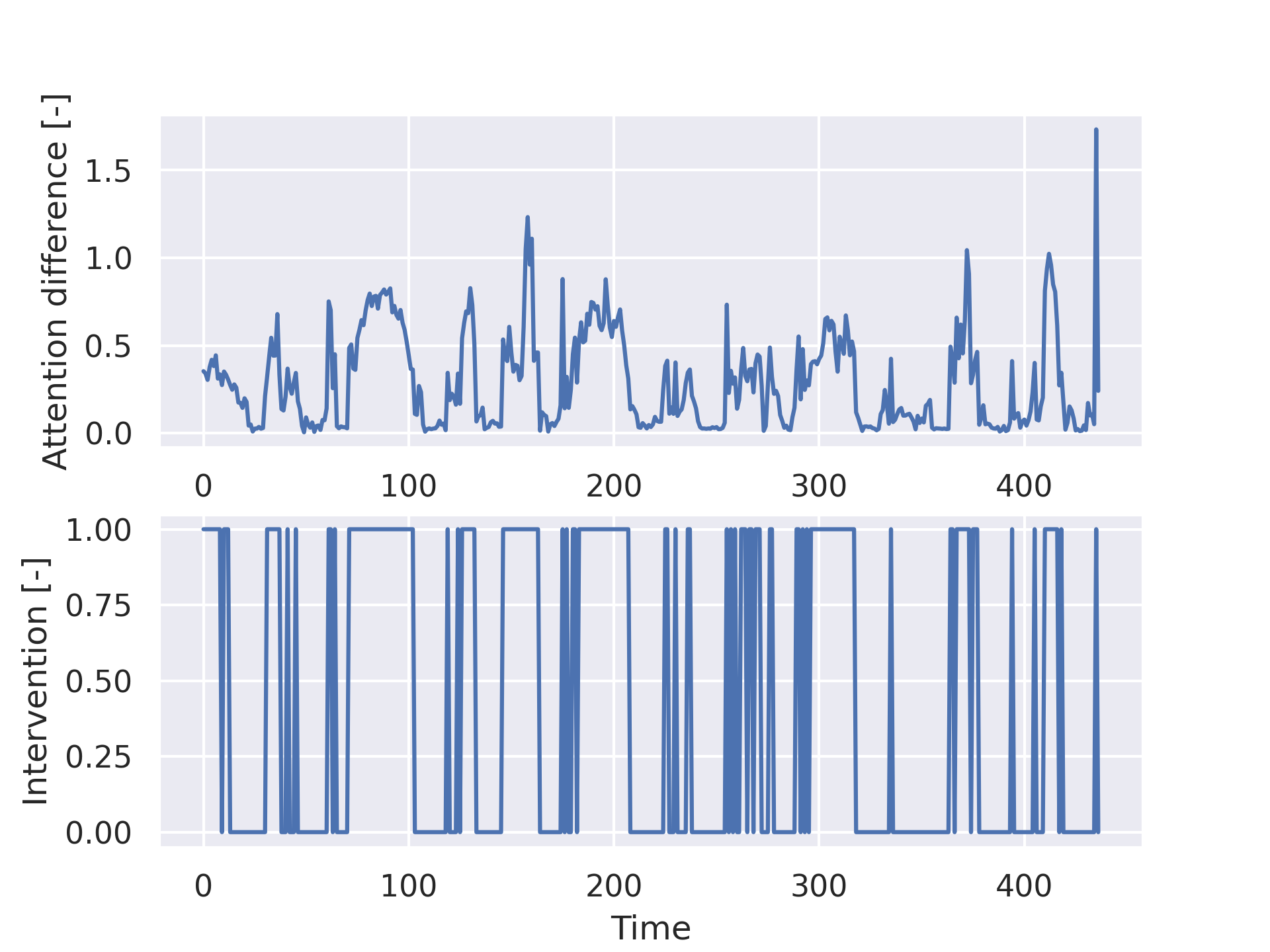}    
	\caption{The intervention process in the field}
	\label{fig: intervention in field}
	\end{center}
\end{figure} 

\begin{table}[ht]
    \centering
    \caption{Experimental results in field test} 
    \begin{tabular}{c|c|c|c|c|c}
    \hline
         Name & Pilot & Guardian& pilot & Guardian& p value  \\
          & Mean & Mean& std & std &   \\
    \hline
         $max(d)\downarrow$ & 1.03     & 0.86 & 0.55  &  0.62 & 0.05 \\
         $max(v)\downarrow$ & 1.56     & 1.10   & 0.52  &  0.50 & 1.57 e-6 \\
    \hline
    \end{tabular}
    \label{tab:fiedlTest}
\end{table}

The result (Fig.~\ref{fig: intervention in field}) shows that the guardian system is able to consistently make interventions when the attention of the pilot and the guardian network do not align. The system can also help fly the drone reaches the target with a lower flight speed and a shorter distance to the optimal flying trajectory (\tabref{tab:fiedlTest}), meaning a smaller risk of hitting other obstacles. The maximum drone speed $max(v)$ with a guardian is smaller than that without a guardian. The maximum distance between the flying trace and the optimal flying trace is defined as $max(d)$. The optimal flying trace is the direct line from the starting point to the goal. $max(d)$ with a guardian is shorter than that without a guardian. The p-value in the table represents cases when the guardian is better than a human. The p-value is less than 0.05, and the difference is significant. The task to navigate to the goal is relatively simple. However, the improvement in performance demonstrated the effectiveness of the proposed system. We also conducted experiments using the eye-tracking system in the field(Fig.~\ref{fig: Field test setup} which can be found in the supplementary. 

\section{Discussion and limitations}
The thresholds of attention difference were selected to keep a balance of pilot control and intervention during experiments. An alternative way is to map the attention difference profile to the visual features when the pilot is distracted, which can be a criterion for intervention. 
Our current solution is built on the trust between the guardian system and the pilot. Flight safety can not be fully guaranteed by our framework if the guardian system does not check the control from the pilot when the attention of the guardian and pilot is aligned. However, integrating the safety-related constraints in Eq.~\ref{eq:MPC continous control} will keep the aircraft in a stable status. One needs to use a network to predict the features of the constraints from the observed image and use them in Eq.~\ref{eq:MPC continous control}. These constraints can handle multiple obstacle situations as a safety module with our proposed guardian system to prevent unsafe actions from the pilot in the future.


\section{Conclusion} \label{section: conclusion and future work }
We proposed a visual-attention-based air-guardian framework using humans' eye-gazing position and saliency map of a guardian network as visual attention. We showed that the proposed method could improve the performance of pilots in both simulation and real-world through cooperation. The air-guardian system makes interventions based on the attention difference between the air-guardian and the pilot. The performance of the pilot is improved when the air-guardian is in place. Our results demonstrate that the proposed cooperative flight control with an eye-tracking system can balance the trade-off between the level of involvement in the flight and the pilots' expertise and attention in the designed simulated flight scenarios and real drone experiments. 


\section*{APPENDIX}
\subsection*{Robustness analysis of cooperative layer}
\label{sec: appendix: robustness analysis}
The robustness of the proposed framework is analyzed based on Lyapunov analysis. The air-guardian control's safety is guaranteed by the following Assumptions:
\begin{assumption} 
\label{assumption: saftey set assumption f_ai }
The system state is in safety set ($y_{k+1}\in \mathbb{S}$) at step $k+1 $, if the previous state $y_{k} \in \mathbb{S}$ and the control input is derived from the trained network $u = f_{G}(X)$.
\end{assumption} 
\begin{assumption} \label{assumption: saftey of u = u_ai + delta u}
There exists $u'$ as a function of $X$ with a Lipschitz constant $c_s>0$ such that the control policy using $u = u_{G} + u' $ can drive the system into the safety set ($y \in \mathbb{S}$).
\end{assumption}
\begin{rem} 
\label{assumption: phi(x) = du/d(x)}
The Visual BackProp is the first order gradient of the network w.r.t. its inputs, $ \phi_{f}(X)= f'(X) $.
\end{rem}

\begin{theorem}
Based on Assumptions \ref{assumption: saftey set assumption f_ai }-\ref{assumption: saftey of u = u_ai + delta u}, the control policy from Eq.~\ref{eq: attention switch distance}-\ref{eq:MPC continous control} drives the system into the safety set, $y \in \mathbb{S}$.
\end{theorem}
\begin{proof}
The control of the reference and the air-guardian networks are given by: $u_{G} = f_{G}(X),~u_{R} = f_{R}(X)$. 

Based on Remark \ref{assumption: phi(x) = du/d(x)},
\begin{equation}\label{eq: f' = du/dx }
\begin{aligned}
    \frac{\Delta u_{R}}{\Delta X} &= f'_{R}(X) = \phi_{f_{R}}\\
     \frac{\Delta u_{G}}{\Delta X} &= f'_{G}(X) = \phi_{f_{G}}
\end{aligned}
\end{equation}
Please note that the reference represents either a neural network that is trained to represent a human operator or a human operator. 
Based on Assumption \ref{assumption: saftey of u = u_ai + delta u}, there exists $u'$ to be the difference of the pilot control and the air-guardian control that bring the system to safety $y \in \mathbb{S}$ and $u_{R} = u= u_{G} + u'$. Therefore, we can compute the following:
\begin{equation} \label{eq: human ai difference}
\begin{aligned}
    f'_{R} - f'_{G} = \frac{\Delta u'}{\Delta X}
\end{aligned}
\end{equation}
Based on Assumption \ref{assumption: saftey of u = u_ai + delta u} and Eq.~\ref{eq: f' = du/dx }, there exist Lipschitz constant $c_s$, and 
\begin{equation} \label{eq: f'_human -f'_ai diff leq }
\begin{aligned}
    \|\phi_{f_{R}} - \phi_{f_{G}}\| = \|f'_{R} - f'_{G}\| \leq c_s,
\end{aligned}
\end{equation}
such that the control policy $u$ from Eq.~\ref{eq: attention switch distance}-\ref{eq:MPC continous control} drives the system into the safety set $y \in \mathbb{S}$. 
\end{proof}

Equation~\ref{eq: f'_human -f'_ai diff leq } indicates that if the difference of the attention maps $\phi_{f_{R}}$ and $ \phi_{f_{G}}$ is smaller than a positive constant, the pilot control $u_{R}$ can achieve the same performance as the $u_{G}$

\begin{assumption} \label{assumption: stability assumption of f_ai}
There exists $u'$ as a function of $X$ with a Lipschitz constant $c_{st}>0$, and $u  \in [u_{G}- u', u_{G} + u']$, $u_{G} = f_{G}(X)$,  such that $u$ controls the system converge, meaning $||x_{k+1}- x_{ref}|| < ||x_{k}- x_{ref}||$.
\end{assumption}

\begin{theorem} 
\label{eq: theorm stability}
Based on Assumption \ref{assumption: stability assumption of f_ai}, the control policy from Eq.~\ref{eq: attention switch distance}-\ref{eq:MPC continous control} drives the system state $\mathbf{x}$ to converge to the reference $\mathbf{x}_{ref}$.
\end{theorem}

\begin{proof}
Denote the scalar function $V: \mathbb{R}^{m \times \mathrm{H_p}} \rightarrow \mathbb{R}$, the cost function at $k+1$ iteration can be written as: 
\begin{equation}\label{eq: lyponouv functions u a_ai}
\begin{aligned}
    V&(\mathbf{u}_{k+1,...,k+1+\mathrm{H_p}}) \\
    &= \sum_{i=k+1}^{i=k+1+\mathrm{H_p}}  
    (\mathbf{x}_i- \mathbf{x}_{ref})^T \mathbf{Q}_i (\mathbf{x}_i - \mathbf{x}_{ref}) 
\end{aligned}
\end{equation}
Please note that the $u$ inside scalar function $V$ is short for normalized $u-u_{ref}$ around the $u_{ref}$ when $x = x_{ref}$. This makes $V(u) = 0$ when $u = 0$.
Based on assumption \ref{assumption: stability assumption of f_ai},  when $I= 1$, 
\begin{equation}
\begin{aligned}
    V(\mathbf{u_{G}}_{k+1,...,k+1+\mathrm{H_p}}) <  V(\mathbf{u_{G}}_{k,...,k+\mathrm{H_p}})
\end{aligned}
\end{equation}
And $\exists u'$, such that  
\begin{equation}\label{eq: lyponouv functions u a_ai + d u}
\begin{aligned}
    &V(\mathbf{u_{G}}_{k+1,...,k+1+\mathrm{H_p}}  \pm u') \\
    = & \sum_{i=k+1}^{i=k+1+\mathrm{H_p}}  
    (\mathbf{x}_i- \mathbf{x}_{ref})^T \mathbf{Q}_i (\mathbf{x}_i - \mathbf{x}_{ref}) \\
     <  & V(\mathbf{u_{G}}_{k,...,k+\mathrm{H_p}}  \pm u')
\end{aligned}
\end{equation}

From Eq.~\ref{eq: human ai difference}, since $u'$ is with a Lipschitz constant $c_{st}$, for $u_{R}  \in [u_{G}- u', u_{G} + u']$, and 
\begin{equation} \label{eq: f'_human -f'_ai diff leq stability}
\begin{aligned}
    \|f'_{R} - f'_{G}\| = \|\phi_{f_{R}} - \phi_{f_{G}}\| = \frac{\Delta u'}{\Delta X} < c_f = \frac{c_{st}}{\Delta X}
\end{aligned}
\end{equation}
the equation $||x_{k+1}- x_{ref}|| < ||x_{k}- x_{ref}||$ holds.

Equation \ref{eq: f'_human -f'_ai diff leq stability} indicates that if the difference of the attention map $\phi_{f_{R}}$ and $ \phi_{f_{G}}$ is smaller than a positive constant, the human control $u_{R} = u = u_{G} \pm u'$ can make system converge as $||x_{k+1}- x_{ref}|| < ||x_{k}- x_{ref}||$. Therefore, Eq.~\ref{eq: lyponouv functions u a_ai} holds when $I= 0$.

\begin{equation}\label{eq: lyponouv functions u a_ai}
\begin{aligned}
    V&(\mathbf{u_{R}}_{k+1,...,k+1+\mathrm{H_p}}) \\
    &= \sum_{i=k+1}^{i=k+1+\mathrm{H_p}}  
    (\mathbf{x}_i- \mathbf{x}_{ref})^T \mathbf{Q}_i (\mathbf{x}_i - \mathbf{x}_{ref}) \\
    & <  V(\mathbf{u_{R}}_{k,...,k+\mathrm{H_p}})
\end{aligned}
\end{equation}

Combining Eq.~\ref{eq: lyponouv functions u a_ai + d u}, and Eq.~\ref{eq: lyponouv functions u a_ai} , the control policy Eq.~\ref{eq: attention switch distance}-\ref{eq:MPC continous control} makes the cost function converge as 
\begin{equation}\label{eq: lyponouv functions V}
\begin{aligned}
    V(\mathbf{u}_{k+1,...,k+1+\mathrm{H_p}})  <&  V(\mathbf{u}_{k,...,k+\mathrm{H_p}}).
\end{aligned}
\end{equation}

The function $V$ has the following properties:

\begin{equation} \label{eq: lyponouv condition1}
    V(\mathbf{u}_{k,...,k+\mathrm{H_p}}) \geq 0 \quad for \quad \mathbf{u}_{k,...,k+\mathrm{H_p}} \neq 0
\end{equation}
\begin{equation} \label{eq: lyponouv condition2}
    V(\mathbf{u}_{k,...,k+\mathrm{H_p}}) = 0 \quad for \quad \mathbf{u}_{k,...,k+\mathrm{H_p}} = 0
\end{equation}
\begin{equation} \label{eq: lyponouv condition3}
    V(\mathbf{u}_{k,...,k+\mathrm{H_p}}) \rightarrow \infty \quad for \quad \mathbf{u}_{k,...,k+\mathrm{H_p}} \rightarrow \infty
\end{equation}

\begin{equation} \label{eq: lyponouv condition4}
\begin{aligned}
    V(\mathbf{u}_{k+1,...,k+1+\mathrm{H_p}}) - &V(\mathbf{u}_{k,...,k+\mathrm{H_p}}) \leq 0 \quad \\
    &for \quad \mathbf{u}_{k,...,k+\mathrm{H_p}} \neq 0
\end{aligned}
\end{equation}
Conditions in Eqs.~(\ref{eq: lyponouv condition1}-\ref{eq: lyponouv condition4}) prove that the proposed cost function of the optimization in Eq.~(\ref{eq: lyponouv functions u a_ai}) is a Lyapunov function and converges to zero with feasible solutions. Thus, the system controlled by the proposed control scheme Eq.~\ref{eq: attention switch distance}-\ref{eq:MPC continous control} is asymptotically stable and converges to its reference. 
\end{proof}

\bibliographystyle{IEEEtran}
\bibliography{main}  

\end{document}